\documentclass{article}
\usepackage{iclr2027,times}
\usepackage{amsmath,amsfonts,bm}

\def\eqref#1{equation~\ref{#1}}
\def\1{\bm{1}}

\DeclareMathAlphabet{\mathsfit}{\encodingdefault}{\sfdefault}{m}{sl}
\SetMathAlphabet{\mathsfit}{bold}{\encodingdefault}{\sfdefault}{bx}{n}

\usepackage{amsmath,amssymb}
\usepackage{algorithm}
\usepackage{algpseudocode}
\usepackage{booktabs}
\usepackage{float}
\usepackage{graphicx}
\usepackage{xcolor}
\usepackage{hyperref}
\usepackage{url}
\usepackage{multirow}

\definecolor{pendingcolor}{RGB}{160,70,0}

\newcommand{\method}{\textsc{HMED}}

\title{Learning from Revision Consequences:\\
Hindsight Meta-Experience Distillation for Self-Improving Agents}
\author{Qianhan Feng\textsuperscript{1}, Zhongzhen Huang\textsuperscript{2}, Yakun Zhu\textsuperscript{2,3}, Xiaofan Zhang*\textsuperscript{2}, Qi Dou*\textsuperscript{1}\\1 The Chinese University of Hong Kong 2 Shanghai Jiao Tong University 3 Shanghai Innovation Institue}
\iclrfinalcopy

\begin{document}
\maketitle

\begin{abstract}
As agents continuously improve by generating and revising Skills, the process that discovers and refines those Skills becomes a learnable object in its own right. Task-Skills directly act on task execution, whereas Meta-Skills govern how agents discover and improve future Skills; their value therefore emerges through the subsequent search processes they induce. Existing approaches improve Meta-Skills from observed raw Skill-search trajectories and branch outcomes. However, branch performance entangles the effects of the initial discovery state and the Meta-Skill revision that generated the search process, making it difficult to characterize what a particular revision actually changed, and pushing updates toward revisions that benefit from favorable states rather than those that improve the process. We introduce \textbf{\method{}} (\textbf{H}indsight \textbf{M}eta-\textbf{E}xperience \textbf{D}istillation), a mechanism for constructing \textbf{Meta-Experience} for self-improving agents. HMED revisits the completed event from which a revision originates and re-executes the incumbent and revised Meta-Skills from the same restored discovery state, so that the changes associated with the revision can be observed under a shared condition. Each comparison is distilled into a Meta-Experience, a structured record that can be reused by future updates, so that even revisions that are not ultimately retained still contribute a learning signal. Across three interactive agent benchmarks and both open-source and closed-source models, HMED consistently improves Skill discovery performance over strong baselines, shifting Meta-Skill learning beyond branch outcomes toward the consequences of changing the improvement process.
\end{abstract}
\section{Introduction}
\label{sec:introduction}

Agents increasingly improve by turning interaction into reusable capabilities.
A household agent that repeatedly has to locate objects can accumulate a Skill
for finding them, so that a lesson learned once is available later without
being rediscovered \citep{shinn2023reflexion,zhao2023expel,wang2023voyager}. As
agents operate in increasingly open environments, however, improvement is no
longer limited to acquiring individual capabilities. An agent must also
improve the process through which it discovers, evaluates, and refines those
capabilities. We refer to the policy governing Skill discovery and evolution
as a \emph{Meta-Skill}, and to the capabilities it produces as
\emph{Task-Skills}. A Task-Skill determines how an agent performs a task and
can usually be evaluated through task outcomes. A Meta-Skill instead shapes
how future capabilities are generated. Its value is realized only through the
search processes it subsequently guides.

Existing approaches improve Meta-Skills using signals derived from Skill-search
trajectories, descendant outcomes, and accumulated evolution histories
\citep{fernando2023promptbreeder,agrawal2025gepa,wang2026metaskill}. These
signals are attached to complete search branches rather than to the
consequences of individual Meta-Skill revisions. A search branch reflects the
combined influence of where the search started and how the improvement policy
guided it. As a result, its final outcome does not directly reveal the
contribution of a particular Meta-Skill revision: a favorable discovery state
may make a weak revision appear effective, while a useful revision may appear
ineffective when evaluated from a difficult starting point. In other words, a
good branch is not necessarily a good improver.

\begin{figure*}[t!]
    \centering
    \includegraphics[width=0.98\linewidth]{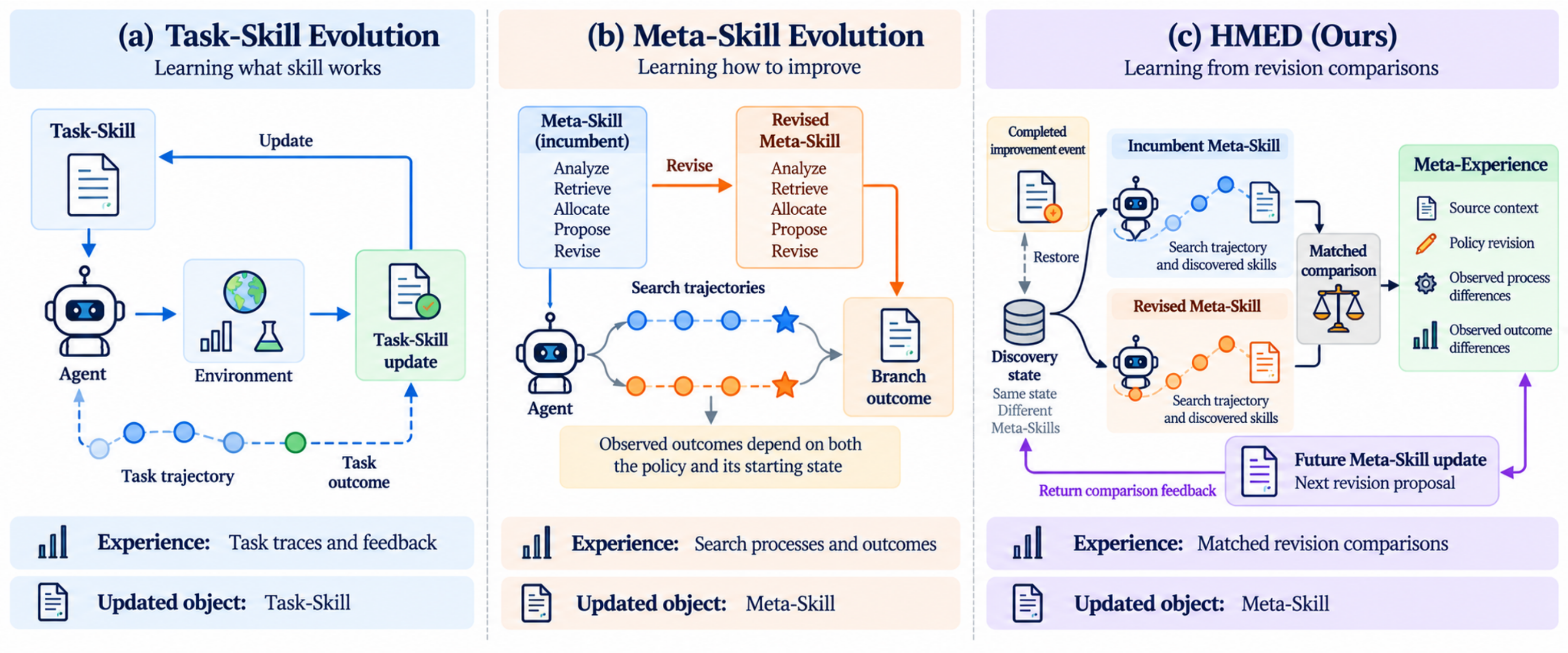}
    \caption{Learning signals in Skill evolution. Task-Skill methods update from
    task outcomes; existing Meta-Skill evolution updates from branch-level
    search signals, where the starting discovery state and the Meta-Skill are
    mixed; HMED updates from matched revision consequences, where the same
    discovery state is restored and only the Meta-Skill is varied.}
    \label{fig:pipelines}
\end{figure*}

This limitation comes from a mismatch between the experience collected by
existing Skill evolution methods and the experience required for learning a
reusable improvement policy. A trajectory records what happened during one
improvement process, whereas it does not characterize what consequences are
associated with a particular revision. What Meta-Skill learning needs is an
experience unit centered on the revision itself: what changed in the
improvement process, and what followed from that change. We call this unit
\emph{Meta-Experience}. Conceptually, it links four aspects of a revision: the
completed event that motivated it, the modification to the Meta-Skill, the
changes observed in the subsequent search process, and the resulting changes
in discovered capabilities. For example, a revision that adjusts the retrieval
component of the Meta-Skill may show which prior experience was consulted
differently and whether that change led to different discoveries. Unlike
Task-level Experience, which describes whether an existing capability can
solve a task, Meta-Experience describes what a change to the improvement
process produced.

We introduce \textbf{Hindsight Meta-Experience Distillation (HMED)} to
construct and reuse this experience. HMED anchors each Meta-Skill revision to
the completed improvement event that motivated it, restores the discovery
state from which that event began, and re-runs the search under both the
incumbent and revised Meta-Skills from that shared starting point. The
resulting paired traces are distilled into a compact Meta-Experience record.
This record enters the next round of Meta-Skill updates, where it can inform
how future revisions are proposed. A comparison therefore serves two roles: it
decides whether to retain the current revision, and it contributes a record
that shapes later revisions. Because the record is kept regardless of whether
the revision is adopted, even a rejected revision leaves a reusable learning
signal.

We evaluate HMED on three interactive agent benchmarks, using DeepSeek-v4-Flash
and Qwen3-32B. HMED achieves the highest mean downstream performance in all
six model--benchmark configurations, and an ALFWorld ablation shows that
returning revision consequences to later updates provides additional gains
over using comparison only for selection. When the learned Meta-Skill is
frozen and the Task-Skill library is reset, it continues to guide effective
discovery at new origins, indicating that HMED learns a reusable improvement
policy rather than fitting the trajectories that produced it. Together, these results suggest that learning an improvement policy benefits from an experience representation aligned with the object being learned. Our contributions are threefold:
\begin{itemize}
\item \textbf{Meta-Experience for Self-Improving Agents.}
We identify the mismatch between branch-level Skill-evolution outcomes and the
revision-level experience required for learning reusable improvement policies,
and introduce Meta-Experience as a distinct experience unit that links the
completed event, the revision, the observed process changes, and the resulting
discovery outcomes of a Meta-Skill update.

\item \textbf{Hindsight Meta-Experience Distillation.}
We introduce HMED, a mechanism that constructs Meta-Experience by grounding
Meta-Skill revisions in completed improvement events, re-running the incumbent
and revised Meta-Skills from matched discovery states, and distilling the
resulting comparison into records that future updates can read.

\item \textbf{Empirical Evidence for Learning from Revision Consequences.}
We evaluate HMED across interactive agent environments and distinguish the
value of consequence feedback from comparison-based selection, showing that
learned Meta-Skills remain effective beyond the trajectories that produced
them.
\end{itemize}
\section{Related Work}
\label{sec:related}

\subsection{Self-Improving Agents and Evolving Improvement Procedures}
\label{sec:rw-selfimprove}

Agents increasingly transform interaction experience into reusable
capabilities. Early work such as Reflexion converts failures into verbal
feedback for future trials \citep{shinn2023reflexion}, ExpeL distills reusable
lessons from repeated interactions \citep{zhao2023expel}, and Voyager maintains
a growing Skill library that can be accumulated, retrieved, and composed
\citep{wang2023voyager}. As the learned object expands from task behavior to
the mechanisms that produce such behavior, self-improvement becomes
increasingly recursive. PromptBreeder evolves the prompts that guide
generation itself \citep{fernando2023promptbreeder}, GEPA updates textual
instructions through reflection on execution trajectories and feedback
\citep{agrawal2025gepa}, and SkillEvolver iteratively refines reusable Skill
artifacts through execution-driven improvement
\citep{zhang2026skillevolver}. Related directions in recursive
self-improvement further explore agents that modify their own reasoning,
search, or capability construction procedures
\citep{zhang2025dgm,kim2026metan,zhang2026hyperagents}. These approaches
progressively shift the focus from improving what an agent does to improving
how an agent improves.

Recent work further advances this direction by making the improvement
procedure itself an editable object. MetaSkill-Evolve introduces a
two-timescale evolution framework in which Task-Skills and Meta-Skills
co-evolve, with the Meta-Skill governing components such as failure analysis,
experience retrieval, budget allocation, candidate proposal, and Skill
revision \citep{wang2026metaskill}. Its updates incorporate accumulated
evolution signals such as descendant improvements and diagnoses, while
maintaining branch-local Skill and Meta-Skill states. This line of work
establishes the improver itself as a persistent object that can be represented,
revised, and inherited. As the scope of learning shifts from task capabilities
to the improvement process, a natural question emerges: what form of
experience should guide the evolution of such an improver?

\subsection{Learning Improvement Policies from Experience}
\label{sec:rw-learnpolicy}

Learning how to improve a search or optimization procedure from previous
experience has a long history in machine learning. Meta-learning studies how
experience across tasks can produce better initializations or adaptation
strategies \citep{finn2017maml,ravi2017optimization}; learned optimizers
parameterize update rules and learn optimization behaviors from prior training
trajectories \citep{andrychowicz2016learning,oh2020discovering}; and
evolutionary computation explores adaptive operator selection and search
strategies based on historical outcomes \citep{lange2023evolving}. These
approaches share a common principle: a search process produces not only a
solution, but also experience that can shape future search. Large language
models extend this idea by representing improvement procedures as
natural-language policies and open-ended agent workflows. EvoPrompt evolves
natural-language prompts through evolutionary operators
\citep{guo2024evoprompt}, while recent agent systems learn from execution
feedback, memory, and accumulated experience to improve future behavior
\citep{zhang2025ace,ouyang2025reasoningbank,wu2025evolver}. As the optimized
object moves from parameterized optimization rules toward language-based
improvement policies that govern diagnosis, retrieval, proposal, and revision,
the experience available to such policies becomes increasingly heterogeneous.
A final search outcome indicates whether a branch succeeded, but it reflects
both the starting discovery state and the improvement procedure that generated
it. HMED instead organizes experience around individual Meta-Skill revisions
within open-ended evolution.

This distinction connects to several neighboring methodological areas.
Hindsight Experience Replay makes failed attempts useful by relabeling the
objective associated with an observed trajectory
\citep{andrychowicz2017hindsight}; in contrast, hindsight in HMED identifies
which Meta-Skill revision should be revisited and which completed improvement
event provides the basis for examining its consequences. Paired comparisons
under shared conditions are widely used to reduce variance and improve
interpretability in simulation and experimental design
\citep{owen2013montecarlo}, while off-policy evaluation estimates the
performance of one policy from observations generated by another
\citep{thomas2015highconfidence,kallus2018instrument,agarwal2014contextual}.
HMED follows a direct comparison perspective by running matched continuations
rather than estimating the value of an unseen policy from logged trajectories.
These directions motivate our study of revision-centered experience for
learning language-based improvement procedures.
\section{Method}
\label{sec:method}

\subsection{Meta-Experience for Meta-Skills}
\label{sec:meta-experience}

An agent that improves itself operates on two levels. At the task level, it
executes a policy $\pi$ on an environment and receives a task outcome
$r \in \mathbb{R}$. At the meta level, it maintains a Meta-Skill $m$ that
governs how Task-Skills are discovered, evaluated, and revised across
windows. A Task-Skill determines how the agent performs a task; a Meta-Skill
instead shapes the capability-production process itself. Let $x$ denote the
discovery state at the start of a window: the Skill library, the visible
history, and the current task context. Running Skill discovery from $x$ under
$m$ produces an observed branch outcome $y = f(x, m)$. The branch outcome is
therefore a function of two arguments: the state the search started from and
the Meta-Skill that guided it.

This distinction creates a core difficulty for Meta-Skill learning: branch
outcomes mix the two arguments of $f$. A revision may appear effective simply
because it was evaluated from a favorable discovery state, and a useful
revision may appear weak because it was evaluated from a difficult one.
Increasing the number of branch observations does not resolve the mixture,
because each branch still corresponds to a different $x$. As a result,
Meta-Skill updates may select revisions that benefit from favorable states
rather than revisions that improve the discovery process.

Meta-Skill learning therefore requires a \emph{revision-centered} form of
experience, one that records the consequences of changing the improvement
process. We call this experience \textbf{Meta-Experience}. Its structure is determined by what is needed to characterize a revision and its consequences: the
completed event that motivated it, the modification it makes, the changes it
produces in the subsequent search process, and the resulting changes in
discovered capabilities. We write the record as
\[
\mathcal{E}_t
=
(e_t,\Delta m_t,
\Delta_{\mathrm{process}},
\Delta_{\mathrm{discovery}}),
\]
where $e_t$ provides the revision's provenance, $\Delta m_t$ describes the
Meta-Skill change, $\Delta_{\mathrm{process}}$ captures how the intermediate
discovery trajectory changed---for example, how prior experience was utilized,
how candidates were generated, and whether a recurring failure was
resolved---and $\Delta_{\mathrm{discovery}}$ records the resulting change in
discovered capabilities. The learning object of Meta-Experience is thus not
task execution itself, but the consequences of changing the improvement
process.

\subsection{Hindsight Meta-Experience Distillation}
\label{sec:hmed}

We propose Hindsight Meta-Experience Distillation (HMED), a mechanism that
constructs Meta-Experience for Meta-Skill learning. HMED revisits a completed
improvement event, restores the discovery state in which the revision was
produced, and re-runs the search under the Meta-Skills before and after the
modification from the same starting point. Here, \emph{hindsight} refers to
revisiting a completed revision after its consequences become observable, and
\emph{distillation} refers to converting the resulting paired traces into a
compact update-level record that later revisions can read. One Meta-Skill
update therefore proceeds in three steps: a revision is proposed and grounded
in a completed event; the incumbent and revised Meta-Skills are re-run from
matched discovery states; and the resulting comparison is distilled into a
Meta-Experience record.

The comparison produces two outputs: a selection signal for whether the revision is retained, and a Meta-Experience record for future updates. A candidate revision is first
produced from the existing improvement process and bound to its source event,
giving the modification a clear problem context and verification location. The
incumbent and revised Meta-Skills are then re-run from matched discovery
states, and the search differences they induce are compared. This comparison is
finally distilled into Meta-Experience and returned to future Meta-Skill
updates. One revision comparison therefore produces two outputs: a selection
signal that determines whether the current revision continues to be used, and a
Meta-Experience record that carries what the comparison revealed about the
proposed change.

\begin{figure*}[t!]
    \centering
    \includegraphics[width=0.98\linewidth]{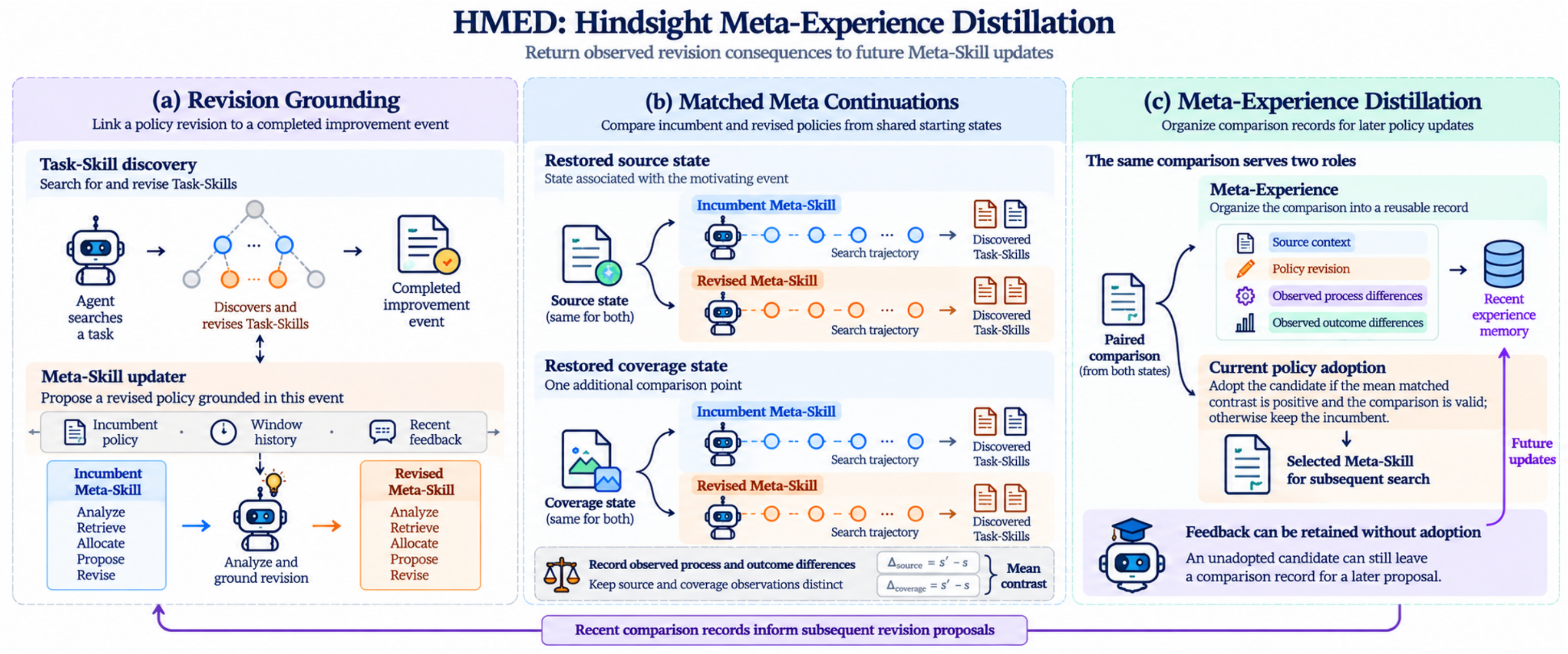}
    \caption{Overview of Hindsight Meta-Experience Distillation (HMED). A
    Meta-Skill revision is anchored to the completed event that motivated it,
    revisited through matched continuations from the same restored state, and
    distilled into a Meta-Experience record that informs both the current
    adoption decision and future revisions.}
    \label{fig:main}
\end{figure*}
\subsection{Revision Grounding in Completed Events}
\label{sec:grounding}

A Meta-Skill revision is the central event around which Meta-Experience is
constructed, and the completed event it originates from provides the
provenance of the resulting record. To make the revision consequence
observable, each revision is bound to a completed Skill discovery attempt. A
Skill discovery attempt refers to a segment of search run by the agent under a
certain discovery state; it may succeed, fail, or be terminated early. When the
agent observes a search bottleneck during this process and proposes a
Meta-Skill modification accordingly, we record that attempt as an event. For
example, a search may fail because the candidate directives do not cover the
current task type; this failure is a discovery event. Grounding revisions in
\emph{completed} events is essential because Meta-Experience describes
observed consequences rather than expectations about a proposed change: an
attempt that has not finished has no search outcome against which the revision
can be evaluated.

Let the current Meta-Skill be $m_t$, the search window be $W_t$, and the
historical Meta-Experience be $\mathcal{E}_{<t}$. The candidate revision and
its source event are represented as
\[
(m'_t,e_t)
=
G(m_t,W_t,\mathcal{E}_{<t}),
\]
where $G$ denotes the Meta-Skill update procedure and $e_t$ denotes the
completed discovery attempt corresponding to this revision. A revision
modifies the Meta-Skill representation while preserving the underlying
discovery framework; it is a textual change to the policy that governs search,
not a replacement of the search procedure itself. The event $e_t$ preserves
the context in which the revision was produced, enabling HMED to return to the
location where the problem first appeared and check whether the modification
changes subsequent search. Given the event, HMED restores its corresponding
discovery state:
\[
x_t^{src}=Restore(e_t).
\]
This state asks: if the revision had been adopted at the location where the
problem arose, would the subsequent discovery process have changed? To observe
whether the revision has a more stable effect, HMED also considers an
additional discovery state:
\[
\mathcal{X}_t=
\{x_t^{src},x_t^{cov}\}.
\]
The source state keeps the comparison consistent with the original problem,
while the additional state supplies a bounded supplementary observation. Event
grounding thus allows the revision to be examined at the location where the
problem first appeared, and the subsequent comparison revolves around the
improvement change that actually occurred. In the implementation, each component of $\mathcal{E}_t$ is instantiated from the corresponding comparison record.

\subsection{Matched Meta Continuations}
\label{sec:matched}

Matched Meta Continuations are the core mechanism by which HMED constructs
Meta-Experience. They hold the discovery state fixed while varying the
Meta-Skill, so that the change associated with the revision can be observed
under a shared reference condition. For example, in the failure scenario from
the previous section, HMED returns to the discovery state at the time of that
failure and re-runs the search with the Meta-Skills before and after the
modification, recording whether the modified version avoids the same failure or
produces a different search trajectory.

For any discovery state $x\in\mathcal{X}_t$, HMED runs the current Meta-Skill
$m_t$ and the candidate revision $m'_t$ separately:
\[
\tau^{-}(x)=\mathcal{S}(x,m_t),
\qquad
\tau^{+}(x)=\mathcal{S}(x,m'_t),
\]
where $\mathcal{S}$ denotes a Skill discovery continuation. The two
continuations start from the same discovery state, so they share the same
initial conditions, and the Meta-Skill is the controlled difference between
them. The difference associated with the revision is therefore
\[
\Delta(x)
=
g(\mathcal{S}(x,m'_t))
-
g(\mathcal{S}(x,m_t)),
\]
where $g(\cdot)$ computes the best-so-far utility improvement along the executed continuation steps. For multiple evaluation states, HMED aggregates:
\[
\Delta_t
=
\frac{1}{|\mathcal{X}_t|}
\sum_{x\in\mathcal{X}_t}
\Delta(x).
\]
Because the discovery state is held fixed, the resulting quantity reflects the
change in subsequent search that is associated with the revision rather than
the advantage of starting from a favorable state. HMED re-runs both
Meta-Skills, so it compares the search difference that actually occurs under
the matched condition rather than a quantity estimated from logged
trajectories.

Based on the matched comparison, the current Meta-Skill is updated according to
the revision consequence:
\[
m_{t+1}
=
\begin{cases}
m'_t,& \Delta_t>0 \ \text{and the comparison passes validity checks},\\
m_t,& \text{otherwise}.
\end{cases}
\]
A comparison is treated as valid when both continuations complete normally, the
candidate text is delivered on every step its own arm realizes, and the
candidate differs textually from the incumbent. Comparisons that fail these
conditions are recorded but not used to update the Meta-Skill.

\subsection{Distilling Revision Consequences into Meta-Experience}
\label{sec:distill}

Distillation is the step that converts the paired traces from a matched
comparison into a compact update-level record. An ordinary execution log
describes what happened during search; Meta-Experience instead captures what
changing the improvement process produced. HMED retains the information required to construct future Meta-Experience records.
The resulting record uses the representation defined in
Section~\ref{sec:meta-experience}, with $e_t$ providing the revision
provenance, $\Delta m_t$ describing the strategy change,
$\Delta_{\mathrm{process}}$ describing how the search process changed, and
$\Delta_{\mathrm{discovery}}$ describing how the capability discovery outcome
changed.

This record enters the future Meta-Skill update process:
\[
(m'_{t+1}, e_{t+1}) = G(m_{t+1}, W_{t+1}, \mathcal{E}_{\leq t})
\]
Historical Meta-Experience gives later revisions information about the
consequences of past modifications, allowing new Meta-Skill updates to
distinguish changes that altered discovery behavior from those that did not. A
rejected revision still contributes a record: it captures what followed from a
direction of change that was tried and found not to improve discovery. HMED retains the consequences of both adopted and rejected revisions, accumulating experience about how discovery procedures evolve.
\section{Experiments}
\label{sec:experiments}

\subsection{Experimental Setup}
\label{sec:setup}

\textbf{Benchmarks.}
We evaluate on three interactive agent benchmarks covering task execution,
open-ended exploration, and scientific discovery.
\textbf{ALFWorld}~\citep{shridhar2021alfworld} provides a text-based embodied
environment in which agents complete everyday tasks through long-horizon
planning, tool use, and environment feedback; task success depends on
accumulating and reusing effective action strategies.
\textbf{ScienceWorld}~\citep{wang2022scienceworld} provides a multi-step
exploration setting in which agents iteratively test hypotheses and refine
strategies, so a useful Skill encodes a procedure rather than a single answer.
\textbf{BiomniBench-DA}~\citep{qu2026biomnibench} targets complex biomedical
analysis tasks that require coordinating multi-step reasoning, tool use, and
knowledge retrieval. Following the official setup, BiomniBench-DA results are
scored by DeepSeek-v4-Flash; the other benchmarks are evaluated using the task
success rates provided by the environment.

\textbf{Models and settings.}
We evaluate on Qwen3-32B~\citep{qwen2025qwen3} and DeepSeek-v4-Flash. All
methods run under identical inference configurations, with reasoning/thinking
modes disabled. Across all experiments, each method interacts only with the
training tasks specified by the benchmark and updates its Task-Skills or
Meta-Skills through its own mechanism; the resulting Skill library and
Meta-Skill are then frozen and evaluated on test tasks not used during
evolution. Unless otherwise stated, results are averaged over three
independent runs.

\begin{table*}[t!]
\caption{Downstream task performance after Skill evolution. ALFWorld and
ScienceWorld report success rate (\%). BiomniBench-DA reports a judge score
out of 100.}
\label{tab:main}
\centering
\small
\setlength{\tabcolsep}{5pt}
\begin{tabular}{@{}llccc@{}}
\toprule
Model & Method & ALFWorld & ScienceWorld & BiomniBench-DA \\
\midrule
\textbf{Qwen3-32B} & No Skill & 45.15 & 16.68 & 14.33 \\
 & Static Skill & 47.76 & 21.55 & 16.36 \\
 & SkillEvolver & 49.25 & 20.59 & 18.33 \\
 & MetaSkill-Evolve & 51.49 & 20.84 & 17.80 \\
 & \method{} (ours) & \textbf{54.11} & \textbf{22.56} & \textbf{20.78} \\
\midrule
\textbf{DeepSeek-v4-Flash} & No Skill & 71.46 & 23.58 & 24.70 \\
 & Static Skill & 73.51 & 26.33 & 22.84 \\
 & SkillEvolver & 76.87 & 24.30 & 25.67 \\
 & MetaSkill-Evolve & 76.49 & 28.70 & 36.80 \\
 & \method{} (ours) & \textbf{80.60} & \textbf{32.47} & \textbf{43.19} \\
\bottomrule
\end{tabular}
\end{table*}

\subsection{Overall Skill Discovery Performance}
\label{sec:main_results}

We evaluate whether \method{} improves downstream task performance after
Skill evolution. We compare with \textbf{SkillEvolver}~\citep{zhang2026skillevolver},
which continually updates Task-Skills, and
\textbf{MetaSkill-Evolve}~\citep{wang2026metaskill}, which additionally allows
the Meta-Skill itself to evolve. We additionally include \textbf{No Skill} and
\textbf{Static Skill} as reference points measuring the contribution of
accumulated Skill knowledge without an evolving mechanism.

Table~\ref{tab:main} reports the results. HMED achieves the highest mean in
all six model--benchmark configurations. Relative to MetaSkill-Evolve, HMED
improves ALFWorld from 51.49 to 54.11 ($+2.62$), ScienceWorld from 20.84 to
22.56 ($+1.72$), and BiomniBench-DA from 17.80 to 20.78 ($+2.98$) on
Qwen3-32B. On DeepSeek-v4-Flash, HMED reaches 80.60 on ALFWorld, 32.47 on
ScienceWorld, and 43.19 on BiomniBench-DA. These results indicate that
organizing revision consequences as Meta-Experience is compatible with
improved downstream performance across different agent backbones and
environments.

Two observations follow from the baseline ordering. First, Static Skill
improves over No Skill on most benchmarks, indicating that accumulated
historical Skills contribute to capability expansion; however, the benefit
depends on whether the current task matches the existing library, as
illustrated by the DeepSeek-v4-Flash BiomniBench-DA cell where Static Skill
falls below No Skill. Second, SkillEvolver and MetaSkill-Evolve are not
uniformly better than Static Skill, and MetaSkill-Evolve does not consistently
outperform SkillEvolver. These comparisons suggest that how improvement
signals are organized and reused can influence downstream self-improvement
performance.

\subsection{Ablation Study on Meta-Experience Construction and Use}
\label{sec:exp-ablation}

The main results show that HMED improves downstream performance. This section
examines which aspects of Meta-Experience construction and use drive that
improvement. Table~\ref{tab:ablation} reports four variants, all retaining
matched comparison.

\begin{table}[!t]
\caption{Ablation of Meta-Experience construction and feedback. All variants
retain matched comparison. Comparison-only removes the explicit feedback
returned to later updates; Outcome-only restricts its content. Values are
success rate (\%) on ALFWorld.}
\label{tab:ablation}
\centering
\small
\setlength{\tabcolsep}{8pt}
\renewcommand{\arraystretch}{1.2}
\begin{tabular}{lcccc}
\toprule
\multirow{2}{*}{Variant} & Event & Matched & Consequence & \multirow{2}{*}{ALFWorld} \\
 & grounding & comparison & feedback & \\
\midrule
\method{} (full) & \checkmark & \checkmark & \checkmark & \textbf{54.11} \\
- Comparison-only & \checkmark & \checkmark & $\times$ & 49.25 \\
- Outcome-only Meta-Experience & \checkmark & \checkmark & Partial & 51.49 \\
- No hindsight grounding & $\times$ & \checkmark & \checkmark & 47.76 \\
\bottomrule
\end{tabular}
\end{table}

\textbf{Selection versus learning.} Matched comparison has two possible uses:
deciding whether to adopt a revision, and informing how future revisions are
proposed. The Comparison-only variant isolates the second by withholding
revision-consequence feedback from later updates while retaining event
grounding, matched continuations, and the adoption rule. According to
Table~\ref{tab:ablation}, the full method exceeds this variant by 4.86 points.
If the benefit of matched comparison were primarily due to selection,
removing consequence feedback would lead to a smaller performance difference.
This comparison isolates the contribution of consequence feedback under matched comparison.

\textbf{Content and source.} Two further variants target the construction of
the record. Removing event grounding reduces the score to 47.76, indicating
that anchoring the comparison to the event that motivated the revision matters
for downstream performance. Restricting the record to outcome-level
information reduces the score to 51.49. Outcome-only remains competitive,
suggesting that outcome information contributes substantially, while the
additional process-level observations provide further benefit in this setting.

\subsection{Analysis of Branch Outcomes and Revision Consequences}
\label{sec:exp-analysis}

The ablation shows which components of Meta-Experience matter. A separate
question is whether the matched comparison merely reproduces the branch
outcome that existing methods rely on. If branch outcome already captured the
consequence of a revision, additional Meta-Experience feedback would provide
limited additional information. For each evaluated revision, we compare the
sign of its original branch outcome with the sign of its matched-continuation
$\Delta$. A revision whose branch outcome is positive but whose matched
$\Delta$ is negative, or vice versa, is a case where branch outcome alone and
matched comparison give opposite directional signals.

\begin{table}[t]
\caption{Agreement between branch outcome and matched-continuation $\Delta$ on
Qwen3-32B. Agreement rate is computed over comparable cases (Agree + Disagree).
Tie indicates that the matched $\Delta$ is zero or the sign cannot be
determined.}
\label{tab:consistency}
\centering
\begin{tabular}{@{}lcccc@{}}
\toprule
Benchmark & Agree & Disagree & Tie & Agreement rate \\
\midrule
ALFWorld & 18 & 15 & 15 & 54.5\% \\
ScienceWorld & 6 & 18 & 12 & 25.0\% \\
\bottomrule
\end{tabular}
\end{table}

Table~\ref{tab:consistency} reports the results. In ALFWorld, 45.5\% of
comparable cases show opposite signs; in ScienceWorld, the disagreement rate
is 75.0\%. These results indicate that branch outcome and
matched-continuation evaluation often provide different directional signals. This
supports treating Meta-Experience as a learning signal that contains
information beyond branch outcome alone. The analysis characterizes the
comparison signal; the feedback ablation above addresses its use in later
updates.

Appendix~\ref{app:case-study} provides a trace showing how a distilled
Meta-Experience record is incorporated into a later update request and
corresponds to a subsequent modification of the Meta-Skill. The trace
illustrates that the record reaches the updater rather than being merely
stored; it does not establish that the record alone caused the subsequent
gain.

\subsection{Reuse of Learned Meta-Skills under New Discovery Origins}
\label{sec:exp-transfer}

The preceding sections evaluate the full evolution pipeline. A separate
question is whether the learned Meta-Skill is a reusable improvement policy or
a fit to the trajectories that produced it. We test this by freezing the
Meta-Skill and resetting the Task-Skill library before new discovery. In this
protocol, we reuse the final Meta-Skill produced by each method on the
training origins; the agent then starts from an initial Skill library on the
held-out set and runs one complete new round of Skill discovery guided by the
frozen Meta-Skill; the resulting Skill library is frozen and evaluated on the
new test set. The held-out and new test sets are obtained by randomly
partitioning the test set into two disjoint halves, neither of which overlaps
with the training tasks; this prevents the reuse phase from adapting directly
to the final evaluation tasks.

\begin{table}[!t]
\caption{Reuse of learned Meta-Skills on unseen discovery origins in ALFWorld, reults reported in success rate (\%) on new set.
The learned Meta-Skill is frozen and the Task-Skill library is reset before
new discovery.}
\label{tab:transfer}
\centering
\small
\setlength{\tabcolsep}{6pt}
\begin{tabular}{@{}lccccc@{}}
\toprule
Model & No Skill & Static Skill & SkillEvolver & MetaSkill-Evolve & \method{} (ours) \\
\midrule
Qwen3-32B & 41.05 & 43.28 & 42.54 & 49.25 & \textbf{50.41} \\
DeepSeek-v4-Flash & 68.65 & 70.15 & 73.13 & 76.12 & \textbf{76.67} \\
\bottomrule
\end{tabular}
\end{table}

Table~\ref{tab:transfer} reports the results. HMED achieves the highest mean
on both base models, with margins of 1.16 and 0.55 points over
MetaSkill-Evolve. Since the Task-Skill library is reset before reuse, the
remaining performance difference provides evidence that the learned Meta-Skill
contributes beyond the stored Skill library. The learned Meta-Skill remains effective under the reuse protocol at new discovery origins, suggesting that HMED captures an improvement strategy that transfers beyond the original discovery trajectory.
\section{Conclusion}
\label{sec:conclusion}

Self-improving agents increasingly accumulate experience from interaction to
discover and revise Skills. As the improvement process itself becomes an
editable object, an important question arises: what form of experience should
guide Meta-Skill learning? We introduce Hindsight Meta-Experience Distillation
(HMED), which shifts the experience unit for Meta-Skill learning from
search-branch outcomes to the consequences of Meta-Skill revisions on
subsequent Skill discovery. HMED anchors each revision to the completed event
that motivated it, compares the incumbent and revised Meta-Skills from matched
discovery states, and distills the resulting revision consequences into a
revision-centered Meta-Experience record. The comparison therefore provides
both an adoption signal for the current revision and a Meta-Experience record
for future updates. Across three interactive agent benchmarks and two base
model families, HMED consistently improves downstream Skill discovery
performance over strong baselines. Further experiments show that its benefit
comes not only from comparison-based selection, but also from the learning
signal provided by Meta-Experience, and that learned Meta-Skills remain
effective when reused under new discovery origins. Together, these results
suggest that once the improvement process becomes an editable object,
these results support organizing experience around revision consequences in self-improving agents.

\bibliography{references}
\bibliographystyle{iclr2027_conference}

\appendix
\newpage
\section{Case Study: Tracing Meta-Experience Through the Update Loop}
\label{app:case-study}

This appendix traces how a distilled Meta-Experience record is incorporated
into later updates rather than merely stored. We analyze one training run of
DeepSeek-v4-Flash on ALFWorld (seed 2). A \emph{Meta-update window} is one
complete cycle of revision proposal, matched evaluation, and updater update;
the run contains four such windows. The case is chosen by position rather than
by outcome: we take the earliest complete chain in the run and include the last
comparison in the same run as a counterpoint. All numbers and texts are quoted
directly from the run records.

Throughout this appendix, \emph{Gain} denotes the best-so-far utility
improvement relative to the starting utility under the same continuation:
$\max_h u_h - u_{\text{start}}$, where the maximum is taken over the executed
continuation steps. \emph{Steps} reports executed steps over allowed steps. Each row in a comparison table corresponds to a single continuation

\subsection{Anadopted Chain: Window 0 to Window 1}

\paragraph{Step 1: The original comparison record.}
Window 0 produced the first Meta-Skill comparison in this run.
Table~\ref{tab:case-w0} lists its key fields.

\begin{table}[!ht]
\caption{Comparison record at window 0. The candidate gains a positive
$\Delta$ at the source state and zero at the coverage state.}
\label{tab:case-w0}
\centering
\small
\begin{tabular}{@{}llcccc@{}}
\toprule
Start & Continuation & $u_{\text{start}}$ & Steps & $u_{\text{best}}$ & Gain \\
\midrule
1 (source) & incumbent $m$ & 0.80 & 2/2 & [0.80, 0.80] & 0.0000 \\
1 (source) & candidate $m'$ & 0.80 & 2/2 & [0.84, 0.84] & 0.0400 \\
2 (coverage) & incumbent $m$ & 0.84 & 2/2 & [0.84, 0.84] & 0.0000 \\
2 (coverage) & candidate $m'$ & 0.84 & 2/2 & [0.84, 0.84] & 0.0000 \\
\bottomrule
\end{tabular}
\end{table}

The outcome is that the candidate isadopted. The record contains both
starting states, the per-arm best-so-far trajectories, the gains, the step
counts, and the inheritance decision. This record becomes the Meta-Experience
that the next window's updater receives.

\paragraph{Step 2: How the next revision uses the Meta-Experience.}
Window 1's candidate rewrites three of the five components: the allocator, the
proposer, and the evolver. The new allocator rule is conditioned on gain values
of the same form as those reported in window 0's record, as shown in
Table~\ref{tab:case-consumption}.

\begin{table}[!ht]
\caption{Correspondence between the gains in window 0's record and the
condition in window 1's allocator rule.}
\label{tab:case-consumption}
\centering
\small
\begin{tabular}{@{}p{0.45\linewidth}p{0.45\linewidth}@{}}
\toprule
Gains in window 0's record & Condition in window 1's allocator rule \\
\midrule
source: $+0.0400$; coverage: $0.0000$ &
``If the previous window produced exactly one archived child with a small
positive $\Delta$ ($\leq 0.05$) and at least one non-archived child, keep $K$
at 2 but require the two children to differ in edit action.'' \\
\bottomrule
\end{tabular}
\end{table}

The phrase ``small positive $\Delta$'' is numerically compatible with the
$+0.0400$ and $0.0000$ in the record. The condition also refers to the archive
status of Task-Skill children, which the paired gains alone do not determine.
We therefore read this as a correspondence between the record's content and a
later rule rather than a derivation of the rule from the record alone. The same
candidate additionally introduces an evolver instruction that prefers
\texttt{modify} over \texttt{add}, with a rationale consistent with the pattern
in the record, where the candidate gained at the source state and was flat at
the coverage state.

\paragraph{Step 3: The next comparison.}
Window 1's own comparison provides a subsequent positive outcome for that
revision, as shown in Table~\ref{tab:case-w1}.

\begin{table}[!ht]
\caption{Comparison result at window 1. The candidate outperforms the incumbent
at both starting states; $\Delta J = +0.04$, and the candidate isadopted.}
\label{tab:case-w1}
\centering
\small
\begin{tabular}{@{}llcccc@{}}
\toprule
Start & Continuation & $u_{\text{start}}$ & Steps & $u_{\text{best}}$ & Gain \\
\midrule
1 (source) & incumbent $m$ & 0.76 & 2/2 & [0.76, 0.76] & 0.0000 \\
1 (source) & candidate $m'$ & 0.76 & 2/2 & [0.80, 0.84] & 0.0600 \\
2 (coverage) & incumbent $m$ & 0.80 & 2/2 & [0.80, 0.84] & 0.0200 \\
2 (coverage) & candidate $m'$ & 0.80 & 2/2 & [0.84, 0.84] & 0.0400 \\
\bottomrule
\end{tabular}
\end{table}

The candidate text is delivered at every step; retrieval and proposals differ
from the incumbent arm at both states, while the failure tag, the allocator
width, and the operation type remain the same. The candidate isadopted and
becomes the incumbent for the following windows.

\paragraph{Step 4: Evidence that the record reached the updater.}
Window 0's comparison is written into a feedback block that appears verbatim in
both of window 1's requests. This establishes that the record was included in
the updater's input; it does not by itself establish that the record caused the
subsequent proposal.

\subsection{A Non-adopted Revision}

The last update in the same run provides a counterpoint.

Window 2's comparison is shown in Table~\ref{tab:case-w2}. The candidate leads
by a wide margin at the source state but loses at the coverage state, resulting
in a tie and keeping the incumbent.

\begin{table}[!ht]
\caption{Comparison record at window 2. The candidate leads at the source state
but falls behind at the coverage state; the outcome is a tie.}
\label{tab:case-w2}
\centering
\small
\begin{tabular}{@{}llcccc@{}}
\toprule
Start & Continuation & $u_{\text{start}}$ & Steps & $u_{\text{best}}$ & Gain \\
\midrule
1 (source) & incumbent $m$ & 0.76 & 2/2 & [0.84, 0.84] & 0.0800 \\
1 (source) & candidate $m'$ & 0.76 & 2/2 & [0.88, 0.88] & 0.1200 \\
2 (coverage) & incumbent $m$ & 0.76 & 1/2 & [0.88] & 0.0600 \\
2 (coverage) & candidate $m'$ & 0.76 & 2/2 & [0.80, 0.84] & 0.0200 \\
\bottomrule
\end{tabular}
\end{table}

Window 3's candidate receives a feedback block containing the comparisons from
windows 1 and 2, and proposes a sharpening of the rules for
\texttt{inventory\_loop} failures. That candidate loses in its own comparison,
as shown in Table~\ref{tab:case-w3}.

\begin{table}[!ht]
\caption{Comparison result at window 3. The candidate ties with the incumbent at
the source state and loses at the coverage state; $\Delta J = -0.03$, and the
candidate is rejected.}
\label{tab:case-w3}
\centering
\small
\begin{tabular}{@{}llcccc@{}}
\toprule
Start & Continuation & $u_{\text{start}}$ & Steps & $u_{\text{best}}$ & Gain \\
\midrule
1 (source) & incumbent $m$ & 0.76 & 2/2 & [0.84, 0.84] & 0.0800 \\
1 (source) & candidate $m'$ & 0.76 & 2/2 & [0.84, 0.84] & 0.0800 \\
2 (coverage) & incumbent $m$ & 0.76 & 1/2 & [0.88] & 0.1200 \\
2 (coverage) & candidate $m'$ & 0.76 & 2/2 & [0.80, 0.84] & 0.0600 \\
\bottomrule
\end{tabular}
\end{table}

The candidate is notadopted, and its changes do not enter the final
Meta-Skill. This counterpoint shows that the feedback channel carries
information rather than a guarantee of success: the updater proposed a new
revision after receiving the record, while the final adoption was still decided
by the matched comparison.

\subsection{Summary of the Case Study}

Together, the two chains illustrate the Meta-Experience loop of HMED:
\begin{itemize}
\item a comparison is recorded and distilled into Meta-Experience;
\item the record is included in a later update request, and part of its content
corresponds to a subsequent proposal;
\item the new proposal is again tested by a matched comparison, and is either
adopted or rejected.
\end{itemize}
The case study provides record-level evidence that Meta-Experience is
incorporated into the update loop rather than merely stored. It illustrates how
the distillation and return of revision consequences enter a later proposal,
without establishing that this return is the sole cause of any downstream
outcome. 

\end{document}